\documentclass[review, 11pt]{elsarticle}

\usepackage{lineno,hyperref}
\modulolinenumbers[5]
\usepackage{framed,multirow}
\usepackage{float}
\usepackage{booktabs}
\usepackage{placeins}
\usepackage[ruled,vlined]{algorithm2e}
\usepackage{setspace}
\usepackage{verbatim}
\usepackage{amssymb}
\usepackage{latexsym}
\usepackage{amsmath}
\usepackage{url}
\usepackage{xcolor}
\usepackage{caption}

\usepackage{algo}
\definecolor{newcolor}{rgb}{.8,.349,.1}

\usepackage{soul}
\usepackage{todonotes}

\journal{Pattern Recognition}

\begin{document}
\begin{frontmatter}

\title{A Novel Bio-Inspired Texture Descriptor based on Biodiversity and Taxonomic Measures}

\author{Steve Tsham Mpinda {Ataky}} 
\ead{steve.ataky@nca.ufma.br}
\author{Alessandro {Lameiras Koerich}\corref{cor1}}
\ead{alessandro.koerich@etsmtl.ca}
\address{\'{E}cole de Technologie Sup\'{e}rieure, Université du Québec\\ 1100, rue Notre-Dame Ouest, H3C 1K3, Montréal, QC, Canada}
\cortext[cor1]{Corresponding author}

\ifpreprint
  \setcounter{page}{1}
\else
  \setcounter{page}{1}
\fi
\begin{abstract}
Texture can be defined as the change of image intensity that forms repetitive patterns resulting from the physical properties of an object's roughness or differences in a reflection on the surface. Considering that texture forms a system of patterns in a non-deterministic way, biodiversity concepts can help texture characterization in images. This paper proposes a novel approach to quantify such a complex system of diverse patterns through species diversity, richness, and taxonomic distinctiveness. The proposed approach considers each image channel as a species ecosystem and computes species diversity and richness as well as taxonomic measures to describe the texture. Furthermore, the proposed approach takes advantage of ecological patterns' invariance characteristics to build a permutation, rotation, and translation invariant descriptor. Experimental results on three datasets of natural texture images and two datasets of histopathological images have shown that the proposed texture descriptor has advantages over several texture descriptors and deep methods.
\end{abstract}

\end{frontmatter}


\section{Introduction}
\label{sec:intro}
Textures are important descriptors used in several image analysis and computer vision applications, such as agriculture \citep{hu2012fish}, recognition of facial expressions \citep{zhao2007dynamic}, object recognition~\citep{khan2009top}, medical image analysis~\citep{ong1996image}, music genre classification~\citep{costa2013music}, remote sensing~\citep{li2015texture}, material~\citep{li2012recognizing} and surface~\cite{Vriesman2019b} recognition, etc. Texture analysis aims at establishing the neighborhood relationship of the texture elements and their position concerning the others (connectivity), the number of elements per spatial unit (density), and their regularity (homogeneity). Texture descriptors developed to characterize image textures by and large fall into statistical methods and geometric methods~\citep{tuceryan1993texture}. The statistical methods aim to discover to which extent some image properties related to its texture may be distributed and derive numerical measures. In contrast, the geometric methods investigate the periodicity in an image and characterize a texture with its relative spectral energy.

Several approaches for texture extraction have been developed in the last three decades, such as gray-level co-occurrence matrix (GLCM)~\citep{haralick1973textural}, Haralick descriptors~\citep{haralick1979statistical}, local binary patterns (LBP)~\citep{pietikainen2011computer}, wavelet transform, 
Markov random fields, 
Gabor texture discriminator, 
local phase quantization, 
local tera pattern, 
binarized statistical image features, 
and fractal models. 
A review of most of these approaches can be found in~\citep{reviewtexture, liu2019bow}.

Even if most of the texture descriptors previously mentioned have proven to be discriminative for texture classification, some do not exploit the color information that may exist in natural and microscopic images, which may bring - to some extent- relevant information. \citet{qi2013exploring} introduced an approach that encodes cross-channel texture correlation and an LBP extension incorporating color information to overcome such limitations. \citet{nsimba2020exploring} have also exploited color information for texture classification. They presented a novel approach to compute information theory measures that capture important textural information from a color image. The accuracy achieved by both approaches is very promising and shows the importance of using color information for texture characterization. 

Researchers have recently been focused on convolutional neural networks (CNNs) due to their effectiveness in object detection and recognition tasks. However, the shape information extracted by CNNs is of minor importance in texture analysis~\citep{texturecnn01}. \citet{texturecnn01} developed a simple texture CNN (T-CNN) architecture for analyzing texture images that pools an energy measure at the last convolution layer and discards the overall shape information analyzed by classic CNNs. Despite the promising results, the trade-off between accuracy and complexity is not so favorable. Other T-CNN architectures have also achieved moderate performance on texture classification \citep{MatosBOK19,fujieda2017wavelet,Vriesman2019}. {Another disadvantage of CNNs is the lack of explainability and interpretability.
\citet{explainability1} argued that interpretability and explainability are two distinct concepts as explainable models are interpretable a priori, but the reverse is not always true. Thus, interpretability alone is insufficient. Therefore, there is a need for explainable models capable of summarizing the reasons for deep learning behavior, gaining the trust about the causes of their decisions.} 

This paper introduces a novel bio-inspired texture (BiT) descriptor based on biodiversity measurements (species richness and evenness) and taxonomic distinctiveness. Ecology primarily exploits these concepts considering patterns in ecosystems. In this work, textural patterns are considered an ecosystem, where both the biodiversity measurements and taxonomic indices are computed and quantified. { \citet{azevedo2020diagnostico} and \citet{de2017lung} exploited distance-based phylogenetic diversity indices as a texture descriptor. \citet{azevedo2020diagnostico} extracted texture features from five different color spaces using some taxonomy indices to detect glaucoma in eye fundus images. Likewise, \citet{de2017lung} extracted texture features employing the same indices for classifying lung cancer images as malignant or benign. Nevertheless, species richness, abundance, and evenness independently were not exploited by both works.
The completeness of biodiversity relies on both richness and taxonomic relation between two organisms. Thus, our approach exploits both sides of ecological diversity indices - as a generalization - whereas the works mentioned above exploited only one side besides focusing on specific problems.}

The BiT descriptor proposed in this paper is a generic descriptor that can characterize texture information on various images. {The BiT descriptor relies on the values of the indexes, which can be explained and interpreted based on the related ecological concepts}. Furthermore, the proposed approach also exploits color information~\citep{qi2013exploring,nsimba2020exploring}. We represent and describe biodiversity as the interaction of pixels with their neighborhood within each image channel (R, G, or B) as well as on a single RGB image. Besides, taxonomic indices and species diversity and richness measures on which the novel BiT descriptor relies are of an underlying use as they capture the all-inclusive behavior of texture image patterns. {They capture the intrinsic properties of the whole ecosystem, although the latter forms a non-deterministic complex system. We state that textural pattern behaves similarly to ecological patterns, in which large populations of units can self-organize into aggregations that generate patterns from processes that are nonlinear and non-deterministic. The complexity, in this case, surges when causality breaks down. The proposed method performs well regardless of this texture nature since biodiversity indices measurements cope with such complexity from the ecosystem perspective.}

The main contribution of this paper is a novel bio-inspired descriptor that exploits species diversity and richness, and taxonomic distinctiveness to build a representation for texture classification. More specifically, the contributions are: (i) modeling each channel of a color image as an ecosystem; (ii) a novel bio-inspired texture (BiT) descriptor combining measurements of species diversity and richness, and taxonomic distinctiveness; (iii) the BiT descriptor is invariant to scale, translation and permutation; (iv) the BiT descriptor is easy to compute and has low computational complexity; (v) the BiT descriptor is a generic texture descriptor that performs well on different image categories, such as natural textures and medical images.

The organization of the remainder of this paper is as follows. Section~\ref{sec:diversity} presents the proposed bio-inspired texture descriptor based on biodiversity measurements and taxonomic distinctiveness. Section~\ref{sec:method} describes a baseline approach to classify texture images, which is used to assess the proposed BiT descriptor's performance and compare its performance with other classical texture descriptors. Section~\ref{sec:experiment} presents the datasets and the experimental protocol. Experimental results, comparison with other texture descriptors and deep approaches, and discussion are presented in Section~\ref{sec:results}. Finally, the conclusions are stated in the last section.

\section{Biodiversity and Taxonomic Distinctiveness}
\label{sec:diversity}
Diversity is a term often used in ecology, and the purpose of diversity indices is to describe the variety of species present in a community or region~\citep{bio5}. A community is defined as a set of species that occurs in a particular place and time. Statistical studies frequently use quantitative measurements of variability, such as mean and variance, while diversity indices describe qualitative variability. Diversity is measured through two variants: (i) species richness, which represents the number of species of a given region; (ii) relative abundance, which refers to the number of individuals of a given species in a given region~\citep{bio2}. However, diversity cannot be measured only in terms of abundance and species richness. It requires the inclusion of a phylogenetic parameter~\citep{clarke1998taxonomic}. Phylogeny is a biology branch responsible for studying the evolutionary relationships between species to determine possible common ancestors. The combination of species abundance with phylogenetic proximity to generate a diversity index is denoted as taxonomic diversity. Taxonomy is the science that deals with classification (creating new taxa), identification (allocation of lineage within species), and nomenclature.

A phylogenetic tree combined with phylogenetic diversity indices compares behavior patterns between species in different areas in biology. Phylogenetic indices (biodiversity and taxonomic indices) can characterize texture due to their potential in describing patterns of a given region/image, regardless of forming a non-deterministic complex system. The richness of details obtained with each indices group is essential for the composition of the descriptors proposed in this paper. We state that these indices are suitable for describing textures due to their ability to analyze the diversity between species in a region.

\subsection{Images as Ecosystems}
\label{sec:ecoimage}
\begin{figure}[!ht]
	\centering
	\includegraphics[width=0.18\textwidth]{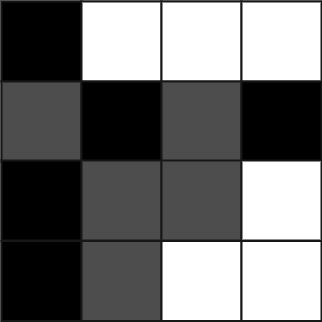}
	\caption[Methodology]{A gray-level image as an abstract model of an ecosystem of three species (three gray levels): white (6 individuals), gray (5 individuals) and black (5 individuals).}
	\label{Fig:imageEcosystem}
\end{figure}

We assume that an image is an abstract model of an ecosystem where: (i) gray levels of pixels in an image correspond to the species in an ecosystem; (ii) pixels in an image correspond to the individuals in an ecosystem; (iii) the number of different gray levels in an image corresponds to species richness in an ecosystem; (iv) the number of distinct gray levels in a specific region of an image corresponds to species abundance in that ecosystem. Another factor is that both the patterns in an ecosystem and the patterns in texture images form a non-deterministic system. For example, Figure~\ref{Fig:imageEcosystem} illustrates an ecosystem with three species, where there are six individuals of white species, five individuals of gray species, and five individuals of black species.

\subsection{Biodiversity and its Measurements}
Biodiversity is defined as the variety within and among life forms on an ecosystem or a site, and it is measured as a combination of richness and evenness across species~\citep{bio2}. Diversity can represent variation in several forms, such as genetic, life form, and functional group. It is worthy of mention that diverse communities are often a sign of fragmented sites where much species richness is contributed by disturbance species~\citep{bio2}. Different objective measures have been proposed as a means to measure biodiversity empirically. The fundamental idea of a diversity index is to quantify biological variability, which, in turn, can be used to compare biological entities composed of direct components, in whether space or time~\citep{bio1}. Biodiversity can be expressed or monitored at different scales and spaces: alpha diversity, beta diversity, and gamma diversity.~\citet{bio4} presents more details on these three types of indices. 

\subsubsection{Diversity Measures}
\label{sec:measurements}
Diversity measurements rely on three assumptions~\cite{bio5}: (i) all species are equal -- richness measurement makes no distinctions among species and considers species exceptionally abundant in the same way as those extremely rare. {In other words, no species is excluded from computing the ecosystem richness due to its abundance}; (ii) all individuals are equal -- there is no distinction between the largest and the smallest individual; however, in practice, the least animals can often escape, for instance, by sampling with nets. This does not necessarily apply to taxonomic and functional diversity measures;; (iii) species abundance is recorded using appropriate and comparable units.

We can translate such assumptions to our abstract model as: (i) {all gray levels are equally taken into account regardless of the number of pixels} -- richness measurement makes no distinctions among gray levels and treats the gray levels that are exceptionally abundant in the same way as those significantly less represented; In other words, all gray levels within an image are taken into account for further calculation, regardless of how non-representative some of them are; (ii) all pixel values are equal -- there is no distinction between the largest and the smallest pixel value; (iii) gray-level abundance has to be recorded using appropriate and comparable units such as the intensity.

Some alpha diversity measures, including measures of richness, dominance, and evenness~\citep{bio9} are described as follows. They represent the diversity within a particular ecosystem: the richness and evenness of individuals within a community. All these indices are computed on a gray-level image of dimensions $m$ and $n$ denoted as $\textbf{I}_{m\times n}$.

\noindent\textbf{Margalef’s} ($\text{d}_{\text{Mg}}$)~\citep{bio6, bio5} and \textbf{Menhinick’s} ($\text{d}_{\text{Mn}}$)~\citep{bio7} \textbf{diversity index} are both the ratio between the number of species ($S$) and the total number of individuals in the sample ($N$):
        \begin{equation}
            \text{d}_{\text{Mg}}= \frac{S-1}{\ln N}
            \label{eq:mg}
        \end{equation}
        
        \begin{equation}
            \text{d}_{\text{Mn}}= \frac{S}{N}
            \label{eq:mn}
        \end{equation}
\noindent{where, $S$ and $N$ denote the number of different gray levels and the total number of pixels in an image, respectively.}
    
\noindent\textbf{Berger-Parker dominance} ($\text{d}_{\text{BP}}$)~\citep{bio8} is the ratio between the number of individuals in the most abundant species ($N_{max}$) and the total number of individuals in the sample:
         \begin{equation}
            \text{d}_{\text{BP}}= \frac{N_{max}}{N}
            \label{eq:bpd}
        \end{equation}
\noindent {where $N_{max}$ denotes {\color{black} the number of pixels belonging to} the most frequent gray level in an image.} 

\noindent\textbf{Fisher’s alpha diversity metric} ($\text{d}_{\text{F}}$)~\citep{bio9,bio121} denotes the number of operational taxonomic units, that is, groups of closely related individuals, and it is defined as:
\begin{equation}
            \text{d}_{\text{F}} = \alpha \ln\left(1 + \frac{N}{\alpha}\right)
            \label{eq:fam}
        \end{equation}
\noindent {where $\alpha$ is approximately equal to the number of gray levels represented by a single pixel.}

\noindent\textbf{Kempton-Taylor index of alpha diversity} ($\text{d}_{\text{KT}}$)~\citep{bio10} measures the interquartile slope of the cumulative abundance curve. $R_1$ and $R_2$ are the 25\% and 75\% quartiles of the cumulative species curve, respectively, $n_r$ is the number of species with abundance $R$, $n_{R_1}$ is the number of individuals in the class where $R_1$ falls, and $n_{R_2}$ is the number of individuals in the class where $R_2$ falls:
\begin{equation}
            \text{d}_{\text{KT}} = \frac{\displaystyle\frac{1}{2}n_{R_1} + \displaystyle \sum_{R_1+ 1}^{R_2 -1} n_r + \frac{1}{2}n_{R_2}}{\log \displaystyle\frac{R_2}{R_1}}
            \label{eq:kt}
        \end{equation}
\noindent {where $n_r$ denotes the number of gray levels with abundance $R$; 
$R_1$ and $R_2$ are the 25\% and 75\% quartiles of the cumulative gray level curve; $n_{R_1}$ is the number of pixels in the class where $R_1$ falls; $n_{R_2}$ is the number of pixels in the class where $R_2$ falls.}

\noindent\textbf{McIntosh’s evenness measure} ($\text{e}_{\text{M}}$)~\citep{bio13} is the ratio between the number of individuals in the $i$-th species and the total number of individuals, and the number of species in the sample:
\begin{equation}
            \text{e}_{\text{M}} = \sqrt{\frac{\displaystyle\sum_{i = 1}^{S}{n_i^2}}{(N-S+1)^2 + S -1}}
            \label{eq:mci}
        \end{equation}
\noindent where $n_i$ denotes the number of pixels of the $i$-th gray level (the summation is over all gray levels).

\noindent\textbf{Shannon-Wiener diversity index} ($\text{d}_{\text{SW}}$)~\citep{bio9} is defined as the proportion of individuals of species $i$ in terms of species abundance ($S$):
\begin{equation}
    \text{d}_{\text{SW}} = - \sum_{i=1}^{S}\left ( p_{i}\: \ln\: p_{i}\right )
    \label{eq:shannon}
\end{equation}

\noindent where $p_i$ denotes the proportion of pixels with the $i$-th gray level. 

\subsection{Taxonomic Indices}
\label{sec:taxonomic}
The ecological diversity indices presented in the previous section are based on the richness and abundance of species present in a community. Nevertheless, such indices may be insensitive to taxonomic differences or similarities. With equal species abundances, they measure but the species richness. Assemblages with the same species richness may either comprise species closely related taxonomically or more distantly related~\citep{bio15}

Taxonomic indices consider the taxonomic relation between different individuals in an ecosystem. The diversity thereof reflects the average taxonomic distance between any two individuals randomly chosen from a sample. The distance can represent the length of the path connecting these two individuals along the phylogenetic tree branches~\citep{bio15}. Taxonomic diversity and taxonomic distinctiveness define the relationship between two organisms randomly chosen in an existing phylogeny in a community \citep{clarke1998taxonomic,bio17}, and three key factors characterize them: (i) number of individuals; (ii) the number of species; (iii) the structure of species connection, that is, the number of edges. Furthermore, \citet{bio17} also proposed the distinctiveness index describing the average taxonomic distance between two randomly chosen individuals through the phylogeny of all species in a sample. This distinctiveness may be represented as taxonomic diversity and taxonomic distinctness~\citep{bio1}, which is described as follows.

\noindent\textbf{Taxonomic diversity} ($\Delta$)~\citep{clarke1998taxonomic} includes aspects of taxonomic relatedness and evenness. In other words, it considers the abundance of species (number of different gray levels) and the taxonomic relationship between them, and whose value represents the average taxonomic distance between any two individuals (pixels), chosen at random from a sample.
    \begin{equation}
        \Delta = \frac{\displaystyle \sum_{i=0}^{S} \sum_{i < j}^{S}w_{ij}x_{i}x_{j}}{\displaystyle\frac{N\left ( N-1 \right )}{2}}
        \label{eq:tdv}
    \end{equation}
\noindent where $x_i$, $x_j$, and $w_{ij}$ represent the number of pixels that have the $i$-th gray level in the image, the number of pixels that have the $j$-th gray level in the image, and the 'distinctness weight' (distance) given to the path length linking pixels $i$ and $j$ in the hierarchical classification, respectively, and $i,j=0,\dots,S$.

\noindent\textbf{Taxonomic distinctiveness} ($\Delta^*$) is a measure of pure taxonomic relatedness. It represents the average taxonomic distance between two individuals (pixels), constrained to pertain to different species (gray levels).
    \begin{equation}
        \Delta^* = \frac{\displaystyle \sum_{i=0}^{S} \sum_{i < j}^{S}w_{ij}x_{i}x_{j}}{\displaystyle \sum \sum_{i< j}^{}x_{i}x_{j}}
        \label{eq:tdt}
    \end{equation}

Different ecological studies, particularly large-scale ones, employ species richness as a measure of biodiversity. Nevertheless, species richness as the sole reflection of biodiversity can present limitations as all species are treated equally without considering phylogenetic relationships. The literature shows that phylogenetic relationships are among the most important factors as they determine species' extinction. Thus, phylogenetic information may be a better indicator of the preservation value than merely the species richness. The studies that verify the distance relationship between the pairs of species are based on a distance matrix computed for all community species. In ecology, this distance matrix relies on either functional or morphological differences~\citep{izsaki1995application}, on the length of the branches of the phylogenetic relationships based on molecular data~\citep{pavoine2005originality}. Accordingly, if the branches' length is unknown, such distances rely on the number of nodes that separate each pair of species~\citep{faith1992conservation}. Therefore, the distance matrix values can be interpreted as the distinctness between each pair of species or between each particular species vis-à-vis all others~\citep{izsaki1995application}. The following indices are based on the distances between pairs of species.

\noindent\textbf{Sum of Phylogenetic Distances} ($\text{s}_\text{PD}$) represents the sum of phylogenetic distances between pairs of species.
\begin{equation}
        \text{s}_\text{PD}  = \left ( \frac{S(S-1)}{2} \right )\frac{\displaystyle \sum \sum_{i < j^2}^{}ij^ai^aj}{\displaystyle \sum \sum_{i<j^a}i^aj^{}}
        \label{eq:spd}
\end{equation}
{\noindent where $i$ and $j$ denote two distinct gray levels, and $a$ is the number of pixels that have such gray levels.}

\noindent\textbf{Average Distance from the Nearest Neighbor} ($ \text{d}_\text{NN}$) \citep{vellend2011measuring} represents the average distance to the nearest taxon\footnote{Taxon is a group of one or more populations of an organism or organisms seen by taxonomists to form a unit. }.
\begin{equation}
        \text{d}_\text{NN} = \sum_{i}^{S}\min\left ( d_{ij},a_i \right )
\end{equation}
\noindent { where $d_{ij} (i, j = 1, \dots , S)$ is the distance between the species (gray levels) $i$ and $j$; $a$ is the abundance of the referred species, and $S$ the total number of species (gray levels).}

\noindent\textbf{Extensive Quadratic Entropy} ($\text{e}_\text{EQ}$) represents the sum of the differences between gray levels.
\begin{equation}
\text{e}_\text{EQ} = \sum_{i\neq j}^{S} d_{ij}
\end{equation}

\noindent\textbf{Intensive Quadratic Entropy} ($\text{e}_\text{IQ}$) represents the number of species and their taxonomic relationships. It aims at establishing a possible link between the diversity indices and the biodiversity measurement indices. Thus, expressing the average taxonomic distance between two species chosen at random, the relationships between them influence the entropy, unlike other diversity indices.
    \begin{equation}
    \text{e}_\text{IQ} = \frac{\displaystyle\sum_{i\neq j}^{S} d_{ij}}{S^{2}}
    \end{equation}

\noindent\textbf{Total Taxonomic Distinctness} ($\text{d}_\text{TT}$): represents the average phylogenetic distinctiveness added across all species (gray levels).
    \begin{equation}
    \text{d}_\text{TT} = \sum i \frac{\displaystyle\sum_{i\neq j}^{S}d_{ij}}{S-1}
    \label{eq:ttd}
    \end{equation}


It is worth noting that Equations~\ref{eq:mg} to~\ref{eq:tdt} are based on species richness, abundance, and evenness, whereas Equations~\ref{eq:spd} to~\ref{eq:ttd} are based on the pairwise distance between pairs of species. All measurements described in Equations~\ref{eq:mg} to~\ref{eq:ttd} can be computed from an image -- in this paper, from each channel of a color image -- and they result in scalar values. {Normalization is required because dynamic ranges of such scalars are related to species richness, abundance, and their relationship within an image or a region of an image, either directly or inversely.} Therefore, these scalars are concatenated
{and normalized within the interval $[0,1]$ using min-max mapping} to form a $d$-dimensional feature vector named the BiT descriptor.

\begin{figure}[!ht]
	\centering
	\includegraphics[width=0.6\textwidth]{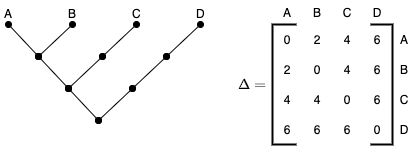}
	\caption[Methodology]{Generic example of a four-species taxonomic tree for four species (A, B, C, and D) and its respective distance matrix. This matrix shows how cumulative branch length, which corresponds to taxonomic distances, is calculated. Image adapted from~\citet{ricotta2004parametric}.}
	\label{Fig:phylo}
\end{figure}

The taxonomic indices require a taxonomic tree to compute species' joint dissimilarity (different gray levels) or pairwise distances between species (different gray levels). The topological distance, defined as the number of edges between two species in the Linnaean taxonomic tree, is the full phylogenetic tree's cumulative branch length. An example of a taxonomic tree and its species distance matrix is shown in Figure~\ref{Fig:phylo}.

\begin{figure}[!ht]
	\centering
	\includegraphics[width=0.70\textwidth]{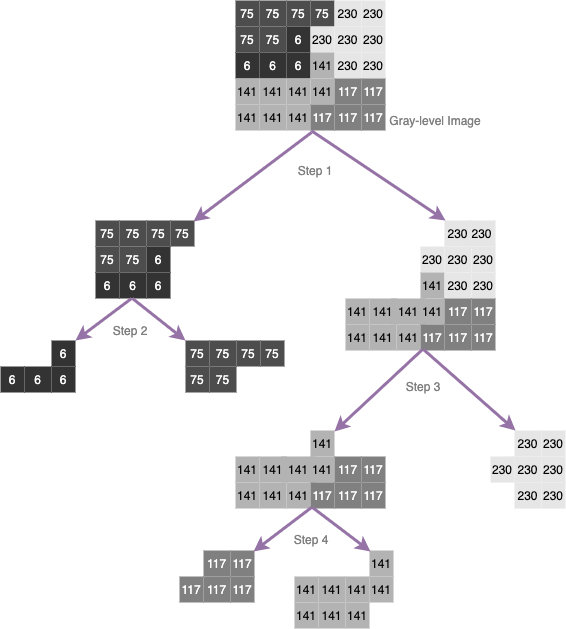}
	\caption[Methodology]{Construction of a phylogenetic tree for computing the taxonomic indexes. In each iteration (step), the image is divided based on species (gray levels). The average species value is used as a threshold at each step.}
	\label{Fig:phylo2}
\end{figure}

Based on the example mentioned above (Figure~\ref{Fig:phylo}), we can derive an instance of the taxonomic tree and its corresponding distance matrix of gray levels (Figure~\ref{Fig:phylo2}). We have represented the taxonomic tree as a matrix, where the distance between two gray levels represents the distance between two species. The division of species in the rooted tree shows the phylogenetic relationship between ancestor species. Such a division allows computing indices connecting diversity, richness, and parenthood between them. Furthermore, a dendrogram can describe the evolutionary relationships between species: the parenthood relationship between gray levels, where the leaves represent the species and the internal nodes represent the common ancestors to the species. This allows establishing an evolutionary connection between the gray levels (species)~\citep{phylo3}, which, in this work, relies on the intrinsic properties of the texture present in an image. Thus, the division of an image or a patch for generating a dendrogram should be based on the parenthood, that is, the similarity between pixels.

\begin{figure}[!ht]

	\centering
		\includegraphics[width=0.35\textwidth]{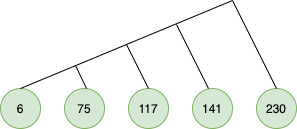}\\[1mm](a)
		\\
		\includegraphics[width=0.45\textwidth]{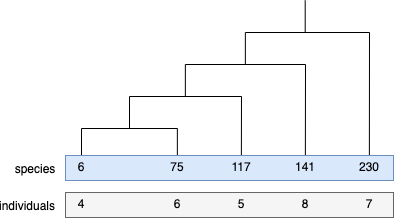}\hspace{1cm}
			\includegraphics[width=0.35\textwidth]{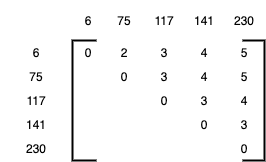}\\[1mm]\hspace{2cm}(b)\hspace{7cm}(c)
	\caption[Methodology]{Example of (a) rooted tree; (b) a dendrogram; (c) and the respective distance matrix of gray levels computed from the image in Figure~\ref{Fig:phylo2}. Note that (a) and (b) are equivalent. The dendrogram allows computing the phylogenetic indexes to infer the phylogenetic relationship between existing gray levels in the original image. Therefrom, the taxonomic indexes are likewise computed.}
	\label{Fig:phylo3}
\end{figure}

Figure~\ref{Fig:phylo2} illustrates the process of division performed in { an image or part of it} to assemble a phylogeny tree (dendrogram) based on the similarity between gray levels for computing the taxonomic indexes. In this case, some iterations are needed to divide the original region/image until a single gray level remains on each leaf. The division is carried out based on a threshold -- {considering all the pixels in the entire image} -- which splits recursively {an image} into two parts, each containing pixels of gray levels above (right) and below (left) the threshold, until the number of species present in each region is 1. Thus, the threshold can be the average gray level of all pixels. From the original image, in the first iteration (step 1), gray levels 6 and 75 (left) are below the threshold ({\color{black}126.57}), whereas gray levels 117, 141, and 230 (right) are above the threshold. The second iteration (step 2) splits the left part resulting from step 1, that is, gray levels 6 (left) and 75 (right), into two parts. Since there are single gray levels in each region resulting from step 2, these regions become leaves. The third iteration (step 3) separates the right part resulting from step 1 into two parts: pixels of gray levels 141 and 117, which are above the threshold ({\color{black}166.15}) go to the left, while pixels of gray-level 230 to the right. Finally, the fourth iteration (step 4) separates the left part from step 3 into two parts: pixels of gray levels 141 and 117. Figure~\ref{Fig:phylo3} illustrates the rooted tree, the dendrogram, and the respective species (gray levels) as well as their characteristics.

\subsection{Properties of BiT Descriptors}
A texture descriptor should have essential properties such as invariance to rotation, translation, and scale for many applications. Furthermore, the descriptor should be easy to calculate. The diversity indices based on species richness measure properties directly related to species, such as their relative abundance and evenness. These measurements are invariant to in-plane rotations and scale (because the true essence of the pattern is invariance). The fundamental idea of diversity indices is to quantify biological variability, which, in turn, can be used to compare biological entities composed of direct components, whether space or time~\citep{bio1}. Biodiversity can be expressed or monitored at different scales and spaces. It is assumed that all species are equal, meaning that richness measurement makes no distinctions among species and treats the species that are exceptionally abundant in the same way as scarce species. All individuals are equal, meaning there is no distinction between the largest and the smallest individual ~\cite{bio5}.

In our abstract model, these assumptions may be expressed as pixels of any gray level are equal. Therefore, the richness measurement makes no distinctions among gray levels and treats pixels that are exceptionally abundant in the same way as extremely less represented pixels. In other words, pixels of all gray levels present in an image are taken into account for further calculation, regardless of how non-representative some are; and all pixel values are equal. There is no distinction between the largest and the smallest pixel value.

In ecology, a pattern is subject to how form remains invariant to changes in measurement. Some patterns retain the same shape after uniformly stretching or shrinking the scale of measurement. The rotational invariance in the ecological pattern has been stated by~\citet{invariance1}, being the most general way to understand commonly observed patterns. From there, species abundance distributions provide a transcendent example, in which the maximum entropy and neutral models can succeed in some cases because they derive from invariance principles. Likewise, as presented by~\citet{invariance2}, diversity is invariant to the species abundance vector's permutation. \citet{invariance3} emphasizes that there is a one-to-one correspondence between abundance vectors and Lorenz curves. Consequently, abundance vectors can be partially ordered according to the Lorenz order, which is permutation-invariant (rotation) and scale-invariant.

Therefore, the BiT descriptor combines statistical and structural approaches and takes advantage of ecological patterns' invariance characteristics to permutation, rotation, and scale by combining species richness, abundance,  evenness, and taxonomic indices.

\subsection{BiT and other Texture Descriptors}
The BiT descriptor shares some characteristics of both GLCM~\citep{haralick1973textural} and LBP~\citep{pietikainen2011computer} descriptors. The BiT descriptor also characterizes textures based on second-order statistical properties, which involves comparing pixels and determining how a pixel at a specific location relates statistically to pixels at different locations.

In ecology, taxonomic indices are approximations of second-order statistics at the species level. These indices are based on group analysis, thus enabling a behavioral exploration of the neighborhood of regions displaced from a reference location. For example, given a distance measurement between pairs of species (pairs of pixels of different gray levels), a classical approach to solving the phylogeny issue can be finding a tree that predicts the observed set of adjoining distances. Such distances are represented in a matrix that indicates the existing phylogenetic distance, reducing it to a simple table of pairwise distances~\citep{bio15,phylo3}.

Furthermore, the BiT descriptor also shares some characteristics of Gabor filters~\citep{gabortext}. Gabor filters explore different periodicities in an image and attempt to characterize a texture at these different periodicities. This analysis is confined to the adjacent neighborhoods of the individual pixels. These within-neighborhood periodicity properties can be used to recognize texture differences between the different regions. Accordingly, phylogenetic trees combined with diversity indices are used in biology to compare behavioral patterns between species in different areas and within-neighborhood. Besides, diversity indices based on species richness are of an underlying use when defining an all-inclusive behavior of an ecosystem, forming a non-deterministic complex system.

\section{Case Study}
\label{sec:method}
\begin{figure}[htpb!]
	\centering
	\includegraphics[width=0.95\textwidth]{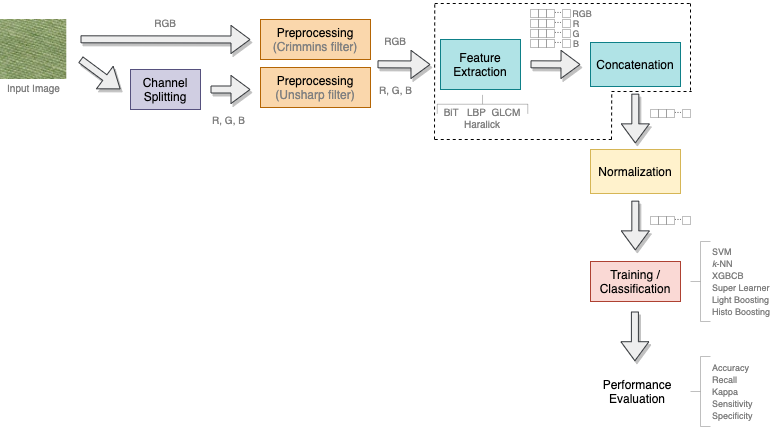}
	\caption[Methodology]{An overview of the proposed scheme to evaluate the BiT descriptor and compare it with other texture extractors.}
	\label{Fig:model}
\end{figure}

This section presents how the proposed bio-inspired texture descriptor can be integrated with image processing and machine learning algorithms for classification tasks. The proposed classification scheme is structured into five stages: image channel splitting, preprocessing, feature extraction, {\color{black}normalization, and training/classification}. Figure~\ref{Fig:model} shows an overview of the proposed scheme. Algorithm~\ref{algo:method} integrates the first three steps, and it receives an RGB image as input and provides a $d$-dimensional feature vector of BiT descriptors. An implementation of this algorithm is available as a Python module\footnote{\color{black}\url{https://github.com/stevetmat/BioInspiredFDesc}. The Python class {\bf BiT}({\it image, b\_feat = True, t\_feat = True, unsharp\_filter = True, crimmins\_filter = True, normalization = False}) generates a 56-dimensional feature vector. The library may be found in \url{https://pypi.org/project/Bitdesc/}.} 
The five stages are described as follows.

\paragraph{Channel Splitting:} 
Besides the original RBG image, each image channel (R, G, B) is considered a separate input. Notwithstanding that the texture descriptors presented in Section~\ref{sec:intro} have shown a discriminative ability to classify texture patterns, their performance on natural and microscopic images may be bounded on the condition that they are applied only to gray-level images. Thus, they do not exploit color information. 
{\citet{reviewer21} and \citet{reviewer22} evaluated the impact of color on texture analysis, and they stated that they are distinct phenomena that should be treated individually. Additionally, comparing texture features extracted from gray-level and color images, it has been put forth that color information improves the accuracy under static illumination conditions. However, employing texture and color in parallel is not as powerful as either color or gray-level texture alone.
The channel splitting aims to capture the textural information of color images based on the principle that most ecosystems work in a cause-effect relationship. Such a relationship implies that when one resource is added or lost, it affects the entire ecosystem. Some of the most marked temporal/spacial fluctuations in species abundances are linked to this cause-effect relationship~\citep{shimadzu2013diversity}. Therefore, we characterize the biodiversity in an image by a set of local descriptors generated from the interaction between a pixel and its neighborhood in each channel (R, G, B) and in the RGB image.}

\paragraph{Preprocessing:} It consists of an unsharp filter to highlight image characteristics and a Crimmins filter to remove speckles \citep{preprocess2}. The unsharp filter is applied to each image channel, and the Crimmins filter is applied to the RGB image to improve their quality for the feature extraction step. 

\paragraph{Feature Extraction and {\color{black}Concatenation} :} After the preprocessing step, the images undergo feature extraction, which looks for informative and discriminative characteristics. Images are represented by several measurements organized in feature vectors. From each image, we extract biodiversity measurements (Equations~\ref{eq:mg} to~\ref{eq:shannon}) and taxonomic indices (Equations~\ref{eq:tdv} to~\ref{eq:ttd}), {\color{black}which are concatenated into a single vector. This process is repeated for all the images of a dataset.}

{\color{black}\paragraph{Normalization:} Before the training step, we first split the feature vectors into training and test sets and perform feature normalization over the training data, where values are normalized to the range $[0,1]$ using the min-max normalization. Afterward, we perform normalization on testing instances using the minimum and maximum values of the training variables. In the case of $k$-fold cross-validation, we used the same procedure by splitting the feature vectors into $k$ folds and computing the min-max pairs in the merged training folds. The training data min-max pairs are then used to normalize both the training and the test fold. This procedure is repeated for each new training/test fold during the cross-validation procedure.

\paragraph{Training/Classification:} The final step of the proposed scheme consists of using a shallow approach where feature vectors are split into training and test partitions to train different classification algorithms, as detailed in Section~\ref{sec:experiment}. Finally, the learned models are evaluated and the results obtained are presented and discussed in Section~\ref{sec:results}.} 

\IncMargin{1em}
\begin{algorithm}[htpb!]
\SetKwData{Left}{left}\SetKwData{This}{this}\SetKwData{Up}{up}
\SetKwFunction{Union}{Union}\SetKwFunction{FindCompress}{FindCompress}
\SetKwInOut{Input}{Input}\SetKwInOut{Output}{Output }
\SetKwInOut{Comment}{Description}
\SetAlgoLined
\Comment{Compute BiT descriptor}
\Input{A RGB image $\textbf{I}_{m\times n\times 3}$}
\Output{A $d$-dimensional feature vector $\bf{x}$}

1. Separate the RGB image $\textbf{I}$ in channels $\textbf{I}^\textrm{R}=\textbf{I}[1\dots n, 1\dots m, 1]$, $\textbf{I}^\textrm{G}=\textbf{I}[1\dots n, 1\dots m, 2]$, $\textbf{I}^\textrm{B}=\textbf{I}[1\dots n, 1\dots m, 3]$\;
2. Convert $\textbf{I}$, $\textbf{I}^\textrm{R}$, $\textbf{I}^\textrm{G}$, and $\textbf{I}^\textrm{B}$ to gray-level images $\textbf{I}^\textrm{g}$, $\textbf{I}^\textrm{Rg}$, $\textbf{I}^\textrm{Gg}$, and $\textbf{I}^\textrm{Bg}$ \;
3. Apply unsharp filter to $\textbf{I}^\textrm{Rg}$, $\textbf{I}^\textrm{Gg}$, and $\textbf{I}^\textrm{Bg}$ \;
4. Apply Crimmins filter to $\textbf{I}^\textrm{g}$ \;
5. Compute biodiversity measurements (Equations 1-7) and taxonomic indices (Equations 8-14) for $\textbf{I}^\textrm{Rg}$, $\textbf{I}^\textrm{Gg}$, $\textbf{I}^\textrm{Bg}$, and $\textbf{I}^\textrm{g}$ \;
6. Concatenate the computed measures and indices into a single vector $\bf{x}$ \;
7. Return $\bf{x}$ \;
\caption{Feature\_Extraction\_Procedure}
\label{algo:method}
\end{algorithm}

\section{Experimental Protocol}
\label{sec:experiment}
This section presents the datasets used to assess the performance of the proposed BiT descriptor, which includes natural texture images and histopathological images (HIs), and the experimental protocol to evaluate the properties of the BiT descriptor and its performance on classification tasks. We compare the BiT descriptor's performance with classical texture descriptors such as LBP, GLCM, and Haralick. It is worthy of mention that our contribution relies on the combination of biodiversity measurements and taxonomic indices to build a discriminative descriptor capable of efficiently classifying textures.
\subsection{Texture Datasets}
\begin{figure}[htpb!]
    	\centering
            \includegraphics[width=6.2in]{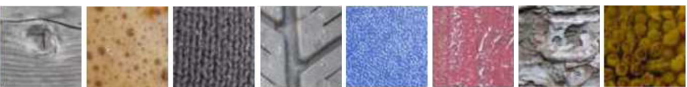}\\[0.5mm](a)
		\\
        \vspace{10pt}
            \includegraphics[width=6.2in]{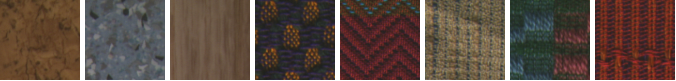}\\[0.5mm](b)
		\\
          \vspace{10pt}
            \includegraphics[width=6.2in]{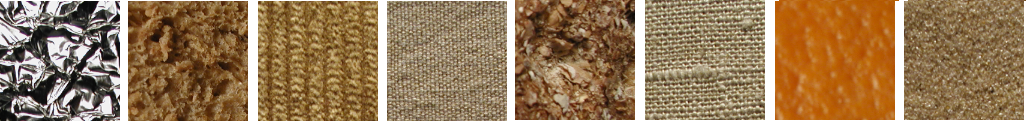}\\[0.5mm](c)
		\\
    	\caption{Samples from the texture datasets: (a) Salzburg, (b) Outex\_TC\_00010\_c, and (c) KTH-TIPS.} 
        \label{fig:texture_datasets}
    \end{figure}

We use three texture datasets that have already been employed for evaluating texture descriptors such as LBP, GLCM, and Haralick~\citep{reviewtexture}. The Salzburg dataset\footnote{\url{http://www.wavelab.at/sources/STex/}} contains 476 color texture images of resolution 128$\times$128, captured around Salzburg in Austria. These images belong to 10 different classes, and 70\% of the images are used for training and validating the classification algorithms, while the remaining 30\% are used for testing. Figure~\ref{fig:texture_datasets}(a) shows some samples from the Salzburg texture dataset.
    
The Outex\_TC\_00010\_c dataset\footnote{\url{http://lagis-vi.univ-lille1.fr/datasets/outex.html}} has a training set consisting of 20 non–rotated color images of each of the 24 classes (480 in total) of illuminant “inca”, color counterpart of the original Outex\_TC\_00010 dataset. The test set consists of 3,840 color images of eight orientations (5, 10, 15, 30, 45, 60, 75, and 90 degrees). Figure~\ref{fig:texture_datasets}(b) shows some samples from the training set of the Outex dataset.

The KTH-TIPS dataset\footnote{\url{https://www.csc.kth.se/cvap/databases/kth-tips/}} contains a collection of 810 color texture images of 200$\times$200 pixels of resolution. The images were captured at nine scales, under three different illumination directions and three different poses, with 81 images per class. Seventy percent of images are used for training, while the remaining 30\% are used for testing. Figure~\ref{fig:texture_datasets}(c) shows some samples from the KTH-TIPS dataset.
    
\subsection{Histopathological Image (HI) Datasets}
HIs are more challenging than pure texture images since HIs usually have other structures such as nuclei (shape) and tissue variations (colors) within the same class. 

\begin{figure}[htpb!]
    	\centering
    	\includegraphics[width=0.55\textwidth]{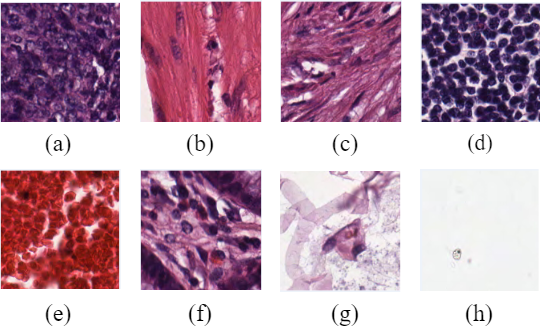}
    	\caption[Methodology]{Samples of the CRC dataset: (a) tumor, (b) stroma, (c) complex, (d) lympho, (e) debris, (f) mucosa, (g) adipose, (h) empty.}
    	\label{Fig:crc}
    \end{figure}

The CRC dataset~\citep{CRC_1} encompasses colorectal cancer histopathology images of dimension 5,000$\times$ 5,000 pixels cropped into 150$\times$150 patches and labeled according to the structure they contain. Eight types of structures are labeled: tumor (T), stroma (ST), complex stroma (C), immune or lymphoid cells (L), debris (D), mucosa (M), adipose (AD), and background or empty (E). Each structure detailed in the CRC dataset has a specific textural characteristic. For example, few shape characteristics are found in cell nuclei formation, which has a rounded shape, but with different coloring due to hematoxylin. The total number of images is 625 per structure type, resulting in 5,000 images. Figure~\ref{Fig:crc} shows samples of each class from the CRC dataset. { The experiments were performed with stratified 5-fold and 10-fold cross-validation.}

The BreakHis dataset~\citep{Spanhol2016} comprises 7,909 microscopic images of breast tumor tissue collected from 82 patients using different magnification factors (40$\times$, 100$\times$, 200$\times$, and 400$\times$). The breast tissues extracted from biopsy usually have some basic structures, such as glands, ducts, and supporting tissue. For example, there will be a difference in textures between an area with a malignant tumor ductal carcinoma and a healthy area. There will be a significant presence of nuclei in the region with carcinoma, identified by the purple color of hematoxylin's reaction with its proteins. The nuclei and many cells in a reduced region make the apparent texture to be noisier. In an area without carcinoma, the epithelial tissue is thin and delimits two regions, lumen and stroma, which have different textural characteristics due to the excess of epithelial cells. The lumen generally presents itself as a homogeneous and whitish region. Due to its reaction to eosin, the stroma shows a pink and homogeneous color, with little noise. At this point, a texture descriptor can assist in detecting carcinomas by characterizing a given texture. 

\begin{figure}[htpb!]
    	\centering
    	\includegraphics[width=0.70\textwidth]{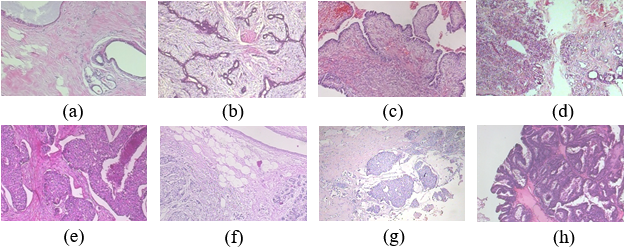}
    	\caption[Methodology]{Example of HIs: (a) Adenosis, (b) Fibroadenoma, (c) Phyllodes, (d) Tabular adenomaa, (e) Ductal carcinoma, (f) Lobular carcinoma, (g) Mucinous carcinoma, (h) Papillary carcinoma, where (a) to (d) are benign and (e) to (f) are malignant tumors.}
    	\label{Fig:breakhis}
    \end{figure}

Nevertheless, the evaluation of types of malignant tumors, that is, differentiation between types of carcinoma on a dataset such as BreaKHis, would need to detect shape to differentiate the papillae from a disorderly cluster of cells. The BreakHis dataset contains 2,480 benign and 5,429 malignant samples (700$\times$460 pixels, 3-channel RGB, 8-bit depth in each channel, PNG format). We used hold-outs with repetition, where 70\% of samples are used for training, and 30\% of samples are used for testing. Figure~\ref{Fig:breakhis} shows samples from each class of the BreakHis dataset. 
    
\subsection{Description of Experiments}
We have carried out three types of experiments to evaluate the proposed BiT descriptor: (i) experiments on texture images in which the accuracy of classification algorithms trained using the BiT descriptor are computed for a comparative analysis with traditional texture descriptors; (ii) experiments on texture images to evaluate invariance of the BiT descriptor to rotation, scale { and intensity}; (iii) experiments on HIs in which measures used frequently in medical imaging such as sensitivity, specificity, and Kappa scores are computed.

The BiT descriptor is evaluated by the accuracy achieved by different classification algorithms on three texture datasets. The same classification algorithms are trained with other texture descriptors, and their performance is compared with the performance achieved with BiT. For a fair comparison with other texture descriptors, we use the same approach described in Section~\ref{sec:method} for all texture descriptors, including the feature extraction procedure described in Algorithm~\ref{algo:method} as well as \textcolor{black}{the preprocessing and normalization steps}. In addition, we have used SVM and $k$-NN and four ensemble learning algorithms: decision tree-based ensemble algorithm that uses a gradient boosting framework (XGBCB), a histogram-based algorithm for building gradient boosting ensembles of decision trees (HistoB), light gradient boosting decision trees (LightB), and super learner (SuperL)~\citep{vanderLaan2007}, which involves the selection of different base classifiers and the evaluation of their performances using a resampling technique. SuperL applies a stacked generalization through out-of-fold predictions during $k$-fold cross-validation. The base classifiers used in SuperL are $k$-NN, decision trees, and ensembles of decision trees such as AdaBoost, bagging, extra trees, and random forest. Furthermore, the $k$-NN was tuned with $k$ values between 1 and 21. The SVM was tuned using a grid search with the hyperparameter $c$ between 0.1 and 2 and linear, polynomial, Gaussian, and sigmoid kernels.

The invariance properties of the proposed BiT descriptor are evaluated on different transformations applied to texture images. We compute the BiT descriptor for each image and compare them to those computed from the transformed images. In this case, feature values should not change with the transformations. {Additionally, we have also evaluated the invariance of the BiT descriptor to monotonic intensity transformations.}  

The BiT descriptor is also evaluated on two HI datasets. In this case, only the classification algorithm that achieved the best performance with the BiT descriptor on the previous experiment is retained. Its performance is compared with the state-of-the-art of these datasets, which includes approaches based on CNNs. The experiments are performed using stratified $k$-fold cross-validation because the related works employed such a strategy.

\begin{table}[H]
\caption{\color{black}Average accuracy (\%) on the test set of Salzburg, Outex, and KTH-TIPS datasets. The overall best result for each dataset is in boldface. The best result for each texture descriptor is marked with $^*$.}
\label{tab:acc_Salz}
\footnotesize
\centering
 \begin{tabular}{r r c c c c c c} 
 \toprule
 & Texture         & \multicolumn{6}{c}{Classification Algorithms}\\ 
\cmidrule{3-8}
Dataset & Descriptors    & XGBCB & HistoB &LightB & SuperL & $k$-NN & SVM\\  
 \midrule
\multirow{3}{*}{Salzburg}
& LBP      &  56.13$\pm$0.024 & 56.89$\pm$0.029 & 57.21$\pm$0.021 & 64.12$^*$ & 32.23 & 62.12 \\
 & GLCM     &  70.74$\pm$0.022 & 68.33$\pm$0.018 & 71.77$\pm$0.016 & 75.04 & 75.38$^*$ & 63.68 \\
 & Haralick &  81.75$\pm$0.015 & 82.27$\pm$0.017 & 84.27$\pm$0.017 & 86.88 & 82.54 & 87.99$^*$ \\
 & BiT      &  {88.65$\pm$0.015} & {88.20$\pm$0.014} & {90.17$\pm$0.013} & \bf{94.23} & 88.65 & {92.33} \\
 
 \midrule
 \multirow{3}{*}{Outex}
& LBP      &  55.30$\pm$0.011 & 57.80$\pm$0.013 & 57.35$\pm$0.014 & 83.21$^*$ & 48.20 & 82.41 \\
 & GLCM     &  94.37$\pm$0.012 & 94.13$\pm$0.006 & 95.52$\pm$0.008$^*$ & 94.29 & 94.37 & 94.06 \\
 & Haralick &  96.15$\pm$0.003 & 96.69$\pm$0.003 & 96.53$\pm$0.004 & 95.5 & 96.92$^*$ & 96.71\\
 & BiT      &  {99.34$\pm$0.006} & {99.68$\pm$0.005} & {98.53$\pm$0.006} & {99.53} & {99.83} & \bf{99.88}\\
 
 \midrule
 \multirow{3}{*}{KTH-TIPS}
 & LBP      &  57.17$\pm$0.021 & 59.51$\pm$0.031 & 57.18$\pm$0.019 & 64.83$^*$ & 58.26 & 61.78 \\
 & GLCM     &  83.53$\pm$0.028 & 83.12$\pm$0.017 & 86.00$\pm$0.022 & 86.83$^*$ & 74.89 & 79.83 \\
 & Haralick &  {90.12$\pm$0.019} & {88.83$\pm$0.019} & {90.94$\pm$0.020} & 93.00 & 89.71 & 94.89$^*$\\
 & BiT      &  92.18$\pm$0.024 & 92.59$\pm$0.022 & {94.65$\pm$0.024} & {95.41} & {95.49} & {\bf 97.87}\\
 
\bottomrule
\multicolumn{8}{l}{{$k$ = 3, 5, and 1 for $k$-NN on Salzburg, Outex, KTH-TIPS datasets, respectively}.}\\
\multicolumn{8}{l}{{Linear kernel and $c$ = 2.0, 1.3, and 1.7 for SVM on Salzburg, Outex, KTH-TIPS datasets, respectively}.}\\
\end{tabular}
\end{table}

\section{Experimental Results and Discussion}
\label{sec:results}

\subsection{Experiments with Texture Datasets}
Table~\ref{tab:acc_Salz} shows the accuracy achieved by monolithic classifiers and ensemble methods trained with four texture descriptors on Salzburg, Outex, and KTH-TIPS datasets. \textcolor{black}{The proposed BiT descriptor achieved the best accuracy for most of the classification algorithms on the Salzburg dataset, and the best result was achieved with BiT+SuperL (94.23\%), which outperformed all texture descriptors. The difference in accuracy achieved by BiT and the second and the third-best texture descriptors (Haralick+SVM and GLCM+$k$-NN) is nearly 6\% and 19\%, respectively. The BiT descriptor also provided the best accuracy on the Outex dataset, and BiT+SVM (99.88\%) achieved the best result. The difference in accuracy achieved by BiT+SVM and the second and the third-best texture descriptors (Haralick+$k$-NN and GLCM+LightB) is nearly 3\% and 5\%, respectively. On the KTH-TIPS dataset, the BiT descriptor provided the best accuracy for all the classification algorithms. The best result was achieved with BiT+SVM {(97.87\%)}. The difference in accuracy achieved by BiT and the second and the third-best texture descriptor (Haralick+SVM and GLCM+SuperL) is nearly 3\% and 11\%.}
{Additionally, the McNemar test\footnote{Significance level of 95\%.} has shown a different proportion of errors on the test set for the three datasets. Therefore, there is a statistically significant difference between the best results of the BiT descriptor and the three other feature descriptors.}

{\color{black}We have also evaluated the importance of the preprocessing step in the final accuracy. Overall, there is no clear advantage of preprocessing texture images because, depending on the dataset and classification algorithm, the preprocessing may improve or harm the accuracy of the feature extraction methods. However, it is worth mentioning that the preprocessing played an important role for GLCM and Haralick descriptors for specific classifiers (mainly $k$-NN), where significant differences (greater than 10\%) were observed between the accuracy achieved on the original dataset (lower) and the preprocessed one (higher). In general, the proposed method does not experience a significant variation of accuracy. For instance, the accuracy of BiT+SVM without preprocessing decreases slightly on Salzburg (93.78\%) and KTH-TIPS (97.57\%) datasets and it increases slightly on the Outex dataset (99.99\%).}

A direct comparison of the results presented in Table~\ref{tab:acc_Salz} with other works that employed the same datasets may not be reasonable due to differences in the experimental protocols. For example, the results reported on the Salzburg dataset omit the subclasses used in the experiments and which samples made up the test set. Several works have also used the Outex dataset for texture classification. For instance,~\citet{kth2} presented an approach based on a rotation-invariant LBP, which achieved an accuracy of 96.26\% with a $k$-NN. Nonetheless, it has the downside of not considering color information and global features. \citet{kth3} presented a rotation-invariant, impulse noise resistant, and illumination-invariant approach based on a local spiking pattern, which achieved an accuracy of 86.12\% with a neural network. Notwithstanding, it is not extended for color textures, and it requires many input parameters.

The KTH-TIPS dataset has also been used to evaluate approaches for texture classification. \citet{kth1} presented an approach based on genetic programming and fusion of HOG and LBP features. Such an approach achieved an accuracy of 91.20\% with a $k$-NN. Nevertheless, it does not consider color information and global features. \citet{kth4} presented statistical binary patterns, which are rotational and noise invariant. Such an approach reached an accuracy of 97.73\%, which is slightly lower than the accuracy achieved by BiT+SVM. In addition to being resolution sensitive, this method presents a high computational complexity. \citet{qi2013exploring} studied the relative variance of texture patterns between different color channels using LBP and Shannon entropy to encode the cross-channel texture correlation. They proposed a multi-scale cross-channel LBP (CCLBP), which is rotation-invariant. The CCLBP computes the LBP descriptors in each channel and three scales and compute co-occurrence statistics before concatenating the extracted features. Such an approach achieved an accuracy of 99.01\% for three scales with an SVM, which is 1.14\% higher than the accuracy achieved by BiT+SVM. Notwithstanding, this method is not invariant to scale.

{In addition to the shallow methods, we have also carried out experiments on the three texture datasets with a tiny T-CNN of 11,900 trainable parameters with and without data augmentation (1$\times$, 2$\times$, 4$\times$, and 6$\times$)~\cite{MatosBOK19, Ataky2020}. The T-CNN achieved the best accuracy of 61.06\%, 70.6\%, and 70.22\% for Salzburg, Outex, and KTH-TIPS datasets, respectively. However, these results are far below the accuracy achieved by BiT+SVM and BiT+SuperL reported in Table~\ref{tab:acc_Salz}.}

{
Finally, we have carried out an empirical evaluation to assess the computational time required by the BiT descriptor for feature extraction and compare it with the other three texture descriptors on the three texture datasets. The average time\footnote{Average time in seconds per image for running a Python implementation of Algorithm 1.} per image, \textcolor{black}{including preprocessing,} is as follows: (i) KTH-TIPS dataset: 27.2 msec, 79.8 msec, 106.2 msec, and 1.39 sec for GLCM, Haralick, LBP, and BiT, respectively; (ii) Outex dataset: 24.4 msec, 42.3 msec, 51.1 msec, and 461.8 msec for GLCM, Haralick, LBP and, BiT, respectively; (iii) Salzburg dataset: 12.6 msec, 18.7 msec, 15.8 msec, and 881 msec for GLCM, Haralick, LBP, and BiT, respectively.} \textcolor{black}{The entire methodology was developed using Microsoft Windows 10 operating system, with the Python programming language
running on an Intel Core i7-8850H CPU @ 2.60 GHz, 64-bit operating system, x64-based processor, and 32 GB of RAM memory.}

\subsection{Invariance of the BiT Descriptor}
\begin{figure}[H]
    	\centering
    	\includegraphics[width=0.80\textwidth]{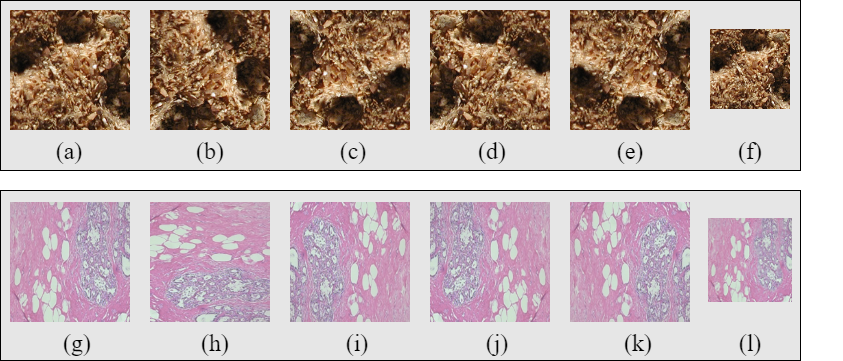}
    	\caption[Methodology]{Example of texture images: (a) original image, (b) rotation 90$^{\circ}$, (c) rotation 180$^{\circ}$, (d) horizontal reflection, (e) vertical reflection, (f) rescaled 50\%. Example of histopathologic images: (g) original image, (h) rotation 90$^{\circ}$, (i) rotation 180$^{\circ}$, (j) horizontal reflection, (k) vertical reflection, (l) rescaled 50\%.}
    	\label{Fig:invariance}
    \end{figure}

Figure~\ref{Fig:invariance} illustrates different transformations applied to texture images (first row) and HIs (second row). We have computed some BiT descriptors for each transformed image, and the non-normalized feature values are presented in Tables~\ref{tab:invariance} and~\ref{tab:invariance2}.

\begin{table}[H]
\caption{Non-normalized feature values computed from different image transformations applied to a texture image (Figure~\ref{Fig:invariance}(a)).}
\label{tab:invariance}
\footnotesize
\centering
 \begin{tabular}{r r r r r r r} 
 \toprule
         & \multicolumn{6}{c}{Transformations}\\ 
\cmidrule{3-7}
 BiT    & \multicolumn{1}{c}{Original} & \multicolumn{2}{c}{Rotation} & \multicolumn{2}{c}{Reflection}  & Rescaling\\
    Features        &  & \multicolumn{1}{c}{90$^o$} & \multicolumn{1}{c}{180$^o$} & \multicolumn{1}{c}{Horizontal} & \multicolumn{1}{c}{Vertical} & \multicolumn{1}{c}{50\%} \\
 \midrule
 $\text{d}_\text{Mg}$           & 2636.49  & 2636.49 & 2636.49 & 2636.49 & 2636.49  & 725.45 \\
  $\text{e}_\text{M}$         & 0.00055  & 0.00055 & 0.00055 & 0.00055 & 0.00055 & 0.00109 \\
 $\text{d}_\text{Mn}$          & 20.3073 & 20.3073  & 20.3073 & 20.3073 & 20.3073 & 10.1634 \\
  $\text{d}_\text{SW}$          & 15.0453  & 15.0453 & 15.0453 & 15.0453 & 15.0453 &  14.9963\\
 $\Delta$     & 101297.7  & 101297.7  & 101297.7 & 101297.7  & 101297.7 & 6253.41 \\
  $\Delta^*$  & 2.002325  & 2.002325  & 2.002325 & 2.002325 & 2.002325 & 2.003482 \\
 $\text{e}_\text{IQ}$  & 2.4900637  & 2.4900637 & 2.4900637 & 2.4900637  & 2.4900637 & 2.4901419 \\
 $\text{d}_\text{NN}$    & 4.9999  & 4.9999 & 4.9999 & 4.9999 & 4.9999 & 4.9999 \\
 \bottomrule
\end{tabular}
\end{table}

\begin{table}[H]
\caption{Non-normalized feature values computed from different image transformations applied to a histopathologic image (Figure~\ref{Fig:invariance}(g)).}
\label{tab:invariance2}
\footnotesize
\centering
 \begin{tabular}{r r r r r r r} 
 \toprule
         & \multicolumn{6}{c}{Transformations}\\ 
\cmidrule{3-7}
 BiT    & \multicolumn{1}{c}{Original} & \multicolumn{2}{c}{Rotation} & \multicolumn{2}{c}{Reflection}  & Rescaling\\
    Features        &  & \multicolumn{1}{c}{90$^o$} & \multicolumn{1}{c}{180$^o$} & \multicolumn{1}{c}{Horizontal} & \multicolumn{1}{c}{Vertical} & \multicolumn{1}{c}{50\%} \\
 \midrule
 $\text{d}_\text{Mg}$           & 1975.95  & 1975.95 & 1975.95 & 1975.95 & 1975.95 & 548.347 \\
 $\text{e}_\text{M}$          & 0.00036  & 0.00036 & 0.00036 & 0.00036 & 0.00036 & 0.00072 \\
 $\text{d}_\text{Mn}$             & 13.2022  & 13.2022 & 13.2022 & 13.2022 & 13.2022 & 6.64831 \\
 $\text{d}_\text{SW}$          & 14.8910  & 14.8910 & 14.8910 & 14.8910 & 14.8910 & 14.6985 \\
 $\Delta$      & 214389.7 & 214389.7& 214389.7& 214389.7& 214389.7& 15287.596 \\
 $\Delta^*$   & 2.00673  & 2.00673 & 2.00673 & 2.00673 & 2.00673 & 2.00710 \\
 $\text{e}_\text{IQ}$      & 2.48115  & 2.48115 & 2.48115 & 2.48115 & 2.48115 & 2.48099 \\
 $\text{d}_\text{NN}$   & 4.9998  & 4.9998 & 4.9998 & 4.9998 & 4.9998 & 4.9998 \\
 \bottomrule
\end{tabular}
\end{table}

The values of BiT descriptors presented in Tables~\ref{tab:invariance} and~\ref{tab:invariance2} show that: (i) all measurements employed are invariant to rotation and reflection as shown in Figures~\ref{Fig:invariance}(a)-(e) and~\ref{Fig:invariance}(g)-(k), since they presented the same values for all texture images or HIs. This also corroborates the fact that BiT descriptors capture the all-inclusive behaviors of patterns in an image; (ii) Shannon-Wiener diversity index ($\text{d}_\text{SW}$), taxonomic distinctness ($\Delta^*$), intensive quadratic entropy ($\text{e}_\text{IQ}$), and the average distance from the nearest neighbor ($\text{d}_\text{NN}$) are invariant to scale as they provided values of the order of other transformations for each of the images. On the other hand, the measures based on richness and abundance show some dependence to scale. By changing the image scale, we somehow affect the proportion of both factors, which affects the resulting values either directly or inversely. However, this may be somehow compensated by normalizing such measures by the total number of pixels. On the other hand, taxonomic indices rely on the parenthood relationship between species. Therefore, they are not affected by the change in scale, as the phylogenetic relationship depends on the intrinsic properties found in the ecosystem (image).

\subsubsection{Invariance to Intensity Changes}
We have also carried out an empirical evaluation to assess the impact of monotonic intensity transformations on the performance of the BiT descriptor. We have applied gamma transformation with values between 0.5 and 3.0 on the test set of Salzburg, Outex, and KTH-TIPS datasets for such an aim. Table~\ref{tab:acc_Salz2} shows the results for selected combinations of the BiT descriptor and classifiers\footnote{The classification models were trained only on the original images.} from Table~\ref{tab:acc_Salz}.

The difference in accuracy between the original and gamma transformed images for the Salzburg dataset is above 5\%. Notwithstanding, the results still slightly higher than other texture descriptors, as shown in Table~\ref{tab:acc_Salz}. On the other hand, the results achieved on the Outex dataset vary less than 1\%, which shows some robustness of the BiT descriptor to intensity changes. There is a difference of nearly 3\% in terms of accuracy between the original and gamma transformed images for the KTH-TIPS dataset.

\begin{table}[htpb!]
\caption{\color{black} Average accuracy (\%) on the three texture datasets with the BiT descriptor applying gamma transformation on the images of the test set.}
\label{tab:acc_Salz2}
\footnotesize
\centering
{ \begin{tabular}{ c c c c c c c c} 
  \toprule
        & Classification & & \multicolumn{5}{c}{Gamma Correction} \\
        \cmidrule{4-8}
Dataset &  Algorithm & Original & 0.5 & 1.5 & 2.0 & 2.5 & 3.0 \\
 \midrule
Salzburg & SuperL & 94.23 & 92.27 & 93.19 & 91.37 & 91.33 & 90.88\\ 
Outex & SVM &  99.88 & 99.45  & 99.61 & 99.45 & 99.76 & 99.69\\
KTH-TIPS &  SVM & 97.87 & 97.65  & 97.23 & 97.65 & 97.35 & 96.41\\ 
  \bottomrule
\end{tabular}
}
\end{table}

{From Table~\ref{tab:acc_Salz2}, the McNemar test has shown similar proportions of errors between the original test set and all transformed test sets of the KTH-TIPS dataset. On the other hand, for the Salzburg dataset, the McNemar test has shown equal proportions of errors with gamma values of 2.0, 2.5, and 3.0, but different proportions of errors with gamma values of 0.5 and 1.5. Finally, for the Outex dataset, the McNemar test has shown similar proportions of errors with gamma values of 0.5, 2.5, and 3.0 and different proportions of errors with gamma values of 1.5 and 2.0. In summary, the proposed BiT descriptor is not invariant to intensity changes, even though it achieved promising results for the KTH-TIPS dataset.}

\begin{table}[htpb!]
\caption{\color{black} Average accuracy (\%) of monolithic classifiers and ensemble methods with the BiT descriptor on the CRC dataset.}
\label{tab:acc_crc}
\footnotesize
\centering
 \begin{tabular}{c c c c c c c} 
 \toprule
Texture         & \multicolumn{6}{c}{Classification Algorithms}\\ 
\cmidrule{2-7}
 Folds    & XGBCB & HistoB & LightB & SuperL & $k$-NN & SVM\\  
  \midrule
5 & 90.80$\pm$0.009 & 89.60$\pm$0.010 & 90.20$\pm$0.013 & {\bf 92.31} & 83.50 & 90.40\\ 
10 & 90.50$\pm$0.010 & 90.71$\pm$0.008 & 90.23$\pm$0.013 & {\bf 92.52} & 83.70 & 90.98\\ 
 \bottomrule
\end{tabular}
\end{table}

\subsection{Experiments with HI datasets}
Table~\ref{tab:acc_crc} shows the accuracy of monolithic classifiers and ensemble methods trained with BiT descriptors on the CRC dataset. Among all classification algorithms, SuperL provided the best results. We have also computed other important metrics used in medical images for BiT+SuperL. Specificity, sensitivity, and Kappa achieved on the CRC dataset are {\color{black}94.19\%, 94.22\%, and 93.53\%}, respectively. {For both cases of Table \ref{tab:acc_crc}, the McNemar test has shown a similar proportion of errors on the test set for the CRC dataset under Super Learner comparing with all the classifiers, except the SVM, which presents a different proportion of errors. Therefore, there is no statistically significant difference in the disagreements between the best results of the BiT descriptor on CRC and all other classifiers. However, there is a statistically significant difference in the disagreements between SVM and Super Learner.}

Table~\ref{tab:acc_crc_relatedWorks} compares the results achieved by BiT+SuperL with the state-of-the-art for the CRC dataset. The proposed descriptor slightly outperforms the accuracy achieved by nearly all other methods. {\color{black}For instance, the difference in accuracy to the second-best method (CNN) is 0.12\%, considering an 8-class classification task for those who used 10-fold cross-validation. In contrast, BiT+SuperL reached the second-best accuracy with a slight difference of 0.29\%} compared with the first-best method (CNN). It is worthy of mention that the success of CNNs relies on the ability to leverage massive labeled datasets to learn high-quality representations. Notwithstanding, data availability for a few fields may be scanty, and therefore CNNs become prohibitive in several domains. The results achieved by the BiT descriptor on the CRC dataset for HI classification have shown that the proposed descriptor works well on other types of images, which have other structures than textures, with no need for data augmentation.

\begin{table}[H]
\caption{\color{black}Average accuracy (\%) of shallow and deep approaches on the CRC dataset.}
\label{tab:acc_crc_relatedWorks}
\footnotesize
\centering
 \begin{tabular}{l c c c} 
  \toprule
  Reference & Approach & 10-fold & 5-fold \\ 
  \midrule
 \citet{CRC_ribeiro2019classification} & Shallow & 97.60$^*$ & --  \\
 \citet{CRC_kather2016multi} & Shallow & 87.40 & --\\
 \citet{CRC_sarkar2017sdl}& Shallow & 73.60 &  -- \\
 {\bf BiT+SuperL} & Shallow & \bf 92.52 & 92.31 \\
 \citet{CRC_wang2017histopathological} & CNN & -- & \bf{92.60} \\
 \citet{CRC_pham2017scaling} & CNN & -- & 84.00 \\
 \citet{CRC_rkaczkowski2019ara} & CNN & 92.40 & 92.20 \\
 \bottomrule
 \multicolumn{4}{l}{$^*$Used 2-classes classification instead.}
\end{tabular}
\end{table}

Table~\ref{tab:acc_break_bal} shows the accuracy of monolithic classifiers and ensemble methods trained with the BiT descriptor on the BreakHis dataset. The SVM classifier achieved the best accuracy for all magnifications, followed by the SuperLearner. Table~\ref{tab:acc_break_precision} shows specificity, sensitivity, and Kappa achieved by BiT+SVM.

\begin{table}[H]
\caption{\color{black}Average accuracy (\%) of classification algorithms with the BiT descriptor on the BreakHis dataset.}
\label{tab:acc_break_bal}
\footnotesize
\centering
 \begin{tabular}{l c c c c} 
  \toprule
  Classification & \multicolumn{4}{c}{Magnification} \\
  \cmidrule{2-5}
 Algorithms &40$\times$ & 100$\times$ & 200$\times$ & 400$\times$\\
    \midrule
 XGBCB    & 94.01 & 94.03 & 92.08 & 91.10\\
 HistoB   & 93.85 & 93.82 & 91.75 & 90.83\\
 LightB   & 94.96 & 94.89 & 93.68 & 92.81\\
 SuperL   & 96.18 & 95.95 & 94.63 & 93.81\\ 
 SVM      & \bf 97.29 & \bf 96.62  & \bf 95.62 & \bf 95.19 \\ 
 \bottomrule
\end{tabular}
\end{table}

\begin{table}[H]
\caption{\color{black} Average specificity, sensitivity, and Kappa (as percentages) for BiT+SVM on the BreakHis dataset.}
\label{tab:acc_break_precision}
\footnotesize
\centering
 \begin{tabular}{c c c c} 
  \toprule
  Magnification & Specificity & Sensitivity & Kappa \\
 \midrule
 40$\times$  & 95.12 & 94.82 & 95.42\\ 
 100$\times$ & 95.35 & 94.43 & 95.21\\
 200$\times$ & 94.05 & 94.18 & 93.22\\
 400$\times$ & 95.12 & 95.24 & 93.72\\
  \bottomrule
\end{tabular}
\end{table}

Table~\ref{tab:acc_breakhis_relatedWorks} compares the results achieved by BiT+SVM with the state-of-the-art for the BreakHis dataset. The proposed descriptor achieved a considerable {\color{black}accuracy of 97.29\%} for 40$\times$ magnification, which slightly outperforms the accuracy of shallow and deep methods. The difference in accuracy between the proposed method and the second-best method (CNN) is {\color{black}about 0.29\%} for 40$\times$ magnification. Notwithstanding, the best CNN method outperforms BiT+SVM for 100$\times$, 200$\times$, and 400$\times$ magnification {\color{black}with a difference of 0.88\%, 1.58\%, and 2.01\%, respectively.} Moreover, Table~\ref{tab:acc_breakhis_relatedWorks} presents the results achieved by \citet{Spanhol2016}, which also used LBP, GLCM, and other texture descriptors with monolithic classifiers and ensemble methods. For instance, the results achieved by BiT+SVM outperform their GLCM approach by {\color{black}22.59\%, 19.82\%, 12.22\% and 13.49\%} for 40$\times$, 100$\times$, 200$\times$ and 400$\times$, respectively.

\begin{table}[H]
\caption{\color{black} Average accuracy (\%) of shallow and deep approaches on the BreakHis dataset. All these works used the same data partition for training and test.}
\label{tab:acc_breakhis_relatedWorks}
\footnotesize
\centering
 \begin{tabular}{l c c c c c} 
 \toprule
  Reference & Method & 40$\times$ & 100$\times$ & 200$\times$ & 400$\times$\\
 \midrule
 \citet{Spanhol2016}$^{*}$ & Shallow &  75.60 & 73.00  & 72.90 & 71.20\\
 \citet{Spanhol2016}$^{+}$ & Shallow &74.70 & 76.80 & 83.40 & 81.70\\
\citet{BreakHis_1}& Shallow &  88.30 & 88.30  & 87.10 & 83.40\\ 
 {\bf BiT+SVM}& Shallow   &  {\bf 97.29} & 96.62  & 95.62 & 95.19\\
 \citet{BrealHis_2}& CNN &  97.00 & {\bf 97.50} & {\bf 97.20} & {\bf 97.20} \\
 \citet{BreakHis_3}& CNN &  92.80 & 93.90  & 93.70 & 92.90 \\
 \citet{BreakHis_4}& CNN &  83.00 & 83.10  & 84.60 & 82.10 \\
 \citet{BreakHis_5}& CNN &  90.00 & 88.40  & 84.60 & 86.10\\
 \bottomrule
 \multicolumn{6}{l}{ $^{*}$: LBP; $^{+}$: GLCM.}
\end{tabular}
\end{table}

Though CNNs have overcome shallow methods for several classification tasks, their advantages on texture images are not so high. {Pre-trained CNN architectures designed for object classification still require fine-tuning some convolutional and fully connected layers on a large amount of data to achieve good performance, including tiny architectures such as T-CNN and T-CNN Inception~\cite{MatosBOK19,Ataky2020}, compact architectures such as EfficientNets and MobileNets\footnote{Number of parameters: T-CNN: 11,900; T-CNN Inception: 1.2M; MobileNetV2: 3.5M; EfficientNetB0: 5.3M}. Besides, CNNs still have explainability and interpretability issues.} In contrast, the amount of data and the computational effort to calculate the BiT descriptor are relatively low. Furthermore, the proposed BiT descriptor is generic, and it does not require retraining or hyperparameter configuration while providing state-of-the-art performance, as the experiments over different datasets have shown. 

\section{Conclusions}
\label{sec:conclusion}
This paper has presented an important contribution to texture characterization using biodiversity measurements and taxonomic distinctiveness. We have proposed a bio-inspired texture descriptor named BiT, based on abstract modeling of an ecosystem as a gray-level image where image pixels correspond to a community of organisms. We have revisited several biodiversity measurements and taxonomic distinctiveness to compute features based on species richness, species abundance, and taxonomic indices. The combination of species richness, species abundance, and taxonomic indices takes advantage of the invariance characteristics of ecological patterns such as reflection, rotation, and scale.

These bio-inspired features form a robust and invariant texture descriptor that can be used together with machine learning algorithms to build powerful classification models. Furthermore, experimental results on texture and HI datasets have shown that the proposed texture descriptor can train different classification algorithms that outperformed traditional texture descriptors and achieved very competitive results compared to deep methods. Therefore, the proposed texture descriptor is promising for mainly dealing with texture analysis and characterization problems. The results demonstrate the promising performance of such a bio-inspired texture descriptor presented.

Considering that the image channels are separated and that the features are extracted using the same measures, it is possible to have redundant and irrelevant features affecting the classification performance. This issue opens the door for a feature selection step. Thus, as future work, we intend to integrate a decision-maker-based multi-objective feature selection into the feature extraction procedure to find a solution that makes a trade-off between the number of features and accuracy. 

\section*{Acknowledgments}
This work was funded by the Natural Sciences and Engineering Research Council of Canada (NSERC) under Grant RGPIN 2016-04855 and by the Regroupement Strategique REPARTI - Fonds de recherche du Quebec - nature et technologie (FRQNT). 

\biboptions{sort&compress}
\bibliographystyle{model1-num-names}
\singlespacing
\bibliography{refs}
\doublespacing
\section*{Supplementary Material}
\noindent All the libraries and implementations are available in the following online public repository:\\ \url{https://github.com/stevetmat/BioInspiredFDesc}. The BiTdesc is also available at: \url{https://pypi.org/project/Bitdesc/}
\end{document}